\documentclass{article}
\usepackage{spconf,amsmath,graphicx,amsfonts}
\usepackage[T1]{fontenc}
\usepackage{titlesec,comment,multirow}
\usepackage{multirow}
\titleformat{\paragraph}
{\normalfont\itshape}{\theparagraph}{1em}{}

\title{A Technical Report: BUT Speech Translation Systems}
\name{Hari Krishna Vydana, Lukas Burget, Jan cernocky \thanks{The work was supported by Czech National Science Foundation (GACR) project ``NEUREM3 No. 19-26934X}}
\address{Brno University of Technology, Faculty of Information Technology, IT4I Centre of Excellence, Czechia}
\begin{document}
\maketitle
\begin{abstract}
The paper describes the BUT's speech translation systems. The systems are English$\longrightarrow$German offline speech translation systems. The systems are based on our previous works~\cite{Jointly_trained_transformers}.
Though End-to-End and cascade~(ASR-MT) spoken language translation~(SLT) systems are reaching comparable performances, a large degradation is observed when translating ASR hypothesis compared to the oracle input text. To reduce this performance degradation, we have jointly-trained ASR and MT modules with ASR objective as an auxiliary loss. Both the networks are connected through the neural hidden representations. This model has an End-to-End differentiable path with respect to the final objective function and also utilizes the ASR objective for better optimization. During the inference both the modules(i.e., ASR and MT) are connected through the hidden representations corresponding to the n-best hypotheses. Ensembling with independently trained ASR and MT models have further improved the performance of the system.

\end{abstract}
\section{Introduction}
\label{sec:intro}
Spoken Language Translation~(SLT) refers to the task of translating the spoken utterance to text in target language. SLT systems which produces the source language text as the intermediate representations and further uses it for generating the translation of the utterance are termed as Cascade SLT systems~\cite{Recent_efforts_Translation,Ney_cascade_SLT}. A cascaded speech translation systems include an automatic speech recognition~(ASR) system followed by a machine translation~(MT) system. SLT systems which does not require the source language text for generating the translation are referred as End-to-End SLT systems~\cite{neural_MT,Listen_and_translate,weiss2017sequence}. End-to-End SLT systems less latent and theoretically simpler models. Cascade systems are superior to End-to-End systems in-terms performance~\cite{IWSLT_2019_campaign}~\cite{sperber2019attention}. 

The SLT systems trained for this work are English-German translation systems. Both the ASR and MT models are trained using transformer~\cite{Transformer} models. In this work, we have also used jointly trained the ASR and MT modules for developing the speech translation system. In our previous work, jointly trained ASR-MT models have worked better than the independently trained ASR+MT with the same data~\cite{Jointly_trained_transformers}. The joint model is ensembled with independently trained ASR and MT models.

\section{Dataset}
The Data for training ASR models is pooled from Librispeech~\cite{Librispeech_dataset}, Must-c and IWSLT data sets~\cite{Must_c_dataset}. The IWSLT train corpus has a lot of noisy utterances and poor alignment quality, the data cleaning strategy specified in ~\cite{inaguma2019multilingual} has been followed. The performances of ASR systems are mentioned in terms of WER. SLT systems used in this work are trained using IWSLT and must-c data sets. The text sentences from the dataset are used to train the MT systems. The performances of ST and MT are reported in terms of Sacre-BLEU. The performances are reported on IWSLT test sets such as dev-2010, tst2010, tst-2013, tst-2014, tst-2015 respectively. For evaluating the SLT systems, the translations corresponding to the segments of audio are concatenated and minimum MWER segmentation~\footnote{https://www-i6.informatik.rwth-aachen.de/web/Software/mwerSegmenter.tar.gz} is used to segment the sentences and BLEU score is evaluated.

\paragraph*{Pre-processing and Feature Extraction:}
All the text is lower-cased and the punctuation symbols are removed. The sentence piece model is trained to obtain a sub-word vocabulary of 5000 tokens for both English and German texts. 80-Dimensional Mel-filter bank along with deltas, delta-deltas are used as features for training the models. The Audio file is segmented using default Kaldi~\cite{Kaldi_toolkit} segmentation, mean and variances of the features are normalized for every segment.  
\section{Transformer ASR}
\label{section_Transformer_ASR}
The ASR systems in this work are trained using transformer models~\cite{Transformer} with characters or sentence piece units as target units~\cite{dong2018_speech_transformer,ESPNET_tranformer}.
The models are trained with 12 encoder and 6 decoder layers. The size of hidden and feed-forward layers are 512, 2048 and the models are trained with 8 heads.The models are trained as mentioned in~\cite{Jointly_trained_transformers}. The models are trained for 60 epochs. A checkpoint of weights are stored for every 20K updates and the final model is obtained by averaging 10-best checkpoints. The CTMs provided by the organizers for the development sets had some time shift in the transcriptions. All the segments that correspond to a ted-talk are concatenated and word error rate~(WER) is computed.
\begin{table}[!ht]
\renewcommand{\arraystretch}{1.5}
\setlength{\tabcolsep}{2pt}
\scriptsize
\begin{center}
\caption{ASR systems trained using Transformer models.}
\label{Transformer_ASR}
\begin{tabular}{cccccc}
\hline
ASR-system &dev-2010 & tst2010 & tst-2013 & tst-2014 & tst-2015\\
\hline
Transformer-BPE&16.33&19.58&17.50&13.54&14.49\\
\hline
Transformer-Char&18.40&21.31&18.97&14.06&15.25\\
\hline
\end{tabular}
\end{center}
\end{table}
The performances of ASR systems are presented in Table~\ref{Transformer_ASR}. The transformer ASR models trained with BPE units as target is referred as Transformer-BPE and the model trained with characters as target units is referred as Transformer-Char. 

\section{Transformer Machine Translation}
\label{section_Transformer_machiene_translation}
Transformer models have been proposed for machine translation in~\cite{Transformer}. Our MT systems are Transformer models with two different input-output granularities i.e., characters-BPE or BPE-BPE. MT-systems are transformers with 6-encoder, 6-decoder layers and feed-forward size of 2048. In Table.~\ref{Transformer_MT}, MT-Trans1, MT-Trans2 are the MT models with both inputs, outputs as BPE units. Similarly, MT-Trans3, MT-Trans4 are the MT models with characters a input and BPE units as outputs. The models referred as MT-Trans1, MT-Trans3 have a hidden size of 256 and the models referred as MT-Trans2, MT-Trans4 has a hidden size of 512. The models are trained as mentioned in~\cite{Jointly_trained_transformers}. The models are decoded with a beam size of 5. The length penalties of characters-BPE and BPE-BPE models are 0.3, 1 respectively. The tokens predicted from the model are converted back to the text and the Sacre BLEU is computed. The performances of the MT systems are presented in Table.~\ref{Transformer_MT}. 

\begin{table}[!ht]
\renewcommand{\arraystretch}{1.5}
\setlength{\tabcolsep}{1pt}
\scriptsize
\begin{center}
\caption{MT systems trained using Transformer models.}
\label{Transformer_MT}
\begin{tabular}{cccccccc}
\hline
Input-Output&MT-system& dev-2010& tst2010& tst-2013& tst-2014& tst-2015\\
\hline
\multirow{2}{*}{BPE-BPE}&MT-Trans1 & 23.53 & 21.83 & 23.56 & 20.61 & 21.67 \\
~&MT-Trans2 & 20.62&21.70&22.37&18.75&20.53\\
\hline
\multirow{2}{*}{CHAR-BPE}&MT-Trans3 & 25.80 & 24.13 & 25.57 & 22.62 & 22.75\\
~&MT-Trans4 & 23.60 & 23.29 & 24.72 & 21.67 & 22.14\\
\hline
\end{tabular}
\end{center}
\end{table}
In Table.~\ref{Transformer_MT}, Column~1 is the MT model and Columns~2-8 are the performances of MT-systems on different test-sets. Columns 2-5 are the performances of MT systems on different development sets provided. It can be observed from Table.~\ref{Transformer_MT} that char-BPE models have performed better than BPE-BPE models. Better performance of Char-BPE models could be attributed to the fact that, during the tokenization of input to the character-BPE models space is used a special character, the encoded space between two words indicates the segmentation of words in a sentence, where it has to learned by the model while using BPE's as input units. Here the sentences that corresponds to the same ted-talk are gathered corpus BLEU of these talks are computed and average corpus BLEU per ted-talk is reported in table~\ref{Transformer_MT}.

\section{Multi-task Training of SLT systems with ASR objective as an Auxiliary Loss}
\label{Joint_ASR_MT_model}
The complete description of the model is presented in~\cite{Jointly_trained_transformers}. As show in the block diagram~\ref{Multi_task_training_with_ASR_auxilary_loss}, the joint model has two modules, An ASR module followed by an MT module. The architectures of ASR and MT modules of the joint model are same as the ones described in sections~\ref{section_Transformer_ASR} and~\ref{section_Transformer_machiene_translation}. The ASR module has gradients corresponding to ASR and MT objectives, while the MT module has gradient corresponding to the MT objective. The MT-module uses the hidden representations from ASR module to train the MT-systems. This architecture has an End-to-End differential pipeline between spoken sequence and the target token sequence, and also utilizes the source transcript for better optimization. The inference of the model is done as described in equation~\eqref{coupling_decoders}.
\begin{figure}[!htbp]
	\begin{flushleft}
	\includegraphics[width=90mm,height=90mm]{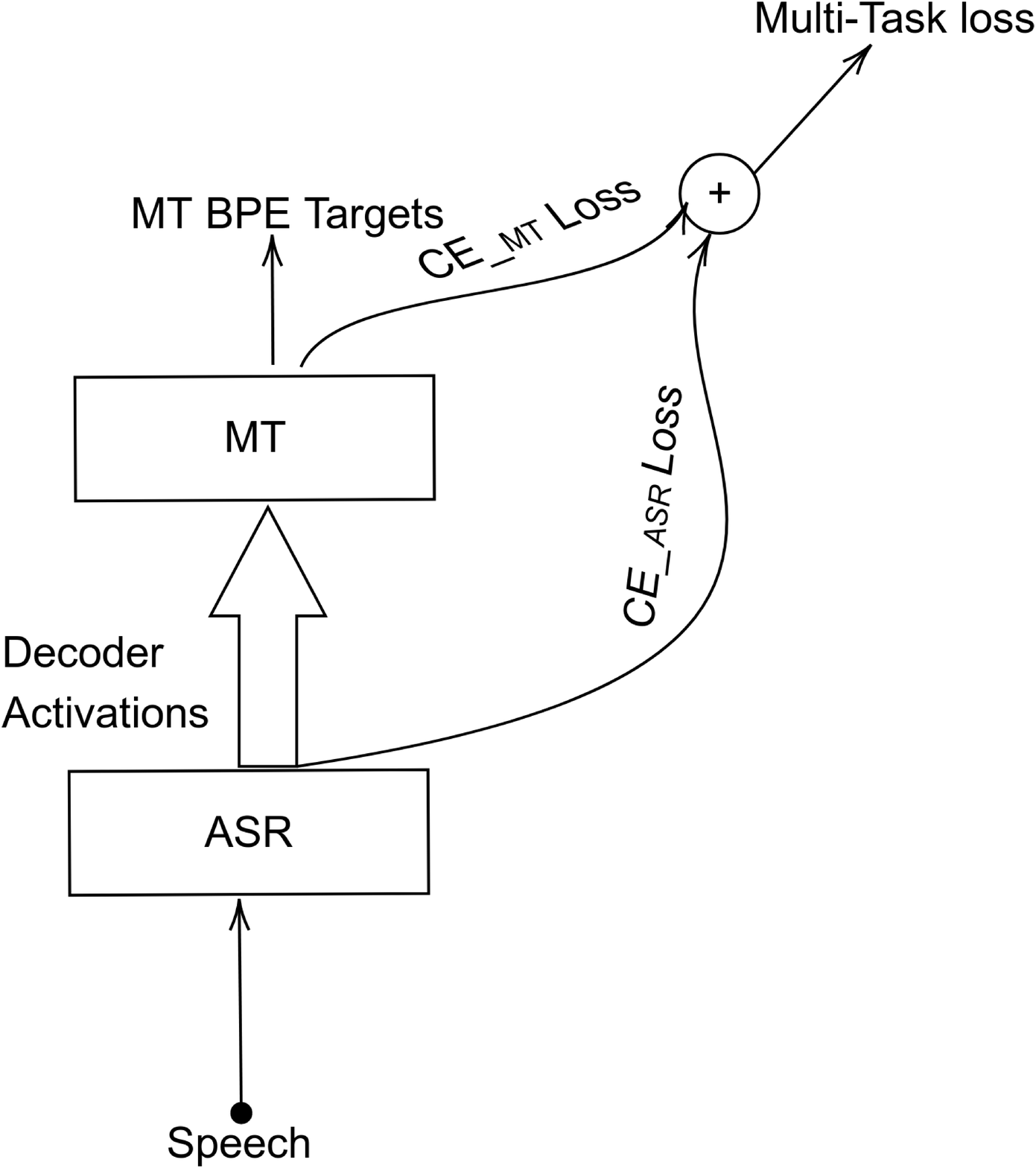}
	\caption{Block diagram of the Transformer based SLT model with ASR objective as auxiliary loss function.}
	\label{Multi_task_training_with_ASR_auxilary_loss}
	\end{flushleft}
\end{figure}

\begin{equation}
\label{coupling_decoders}
\begin{aligned}
P(y|x) & =\arg\max_{y}\sum_z P(y|z)P(z|x))\\
&\approx \arg\max_{y \in \hat{Y}(z),z \in \hat{Z}(x)}log P(y|z) + log P(z|x)
\end{aligned}
\end{equation}

In equation~\eqref{coupling_decoders}, $x$, $y$ and $z$ are source speech, target token sequence and the source token sequence. N-best hypotheses from ASR and MT modules are $\hat{Z}$, $\hat{Y}$. During the inference, ASR Decoder produces the n-best source token sequence and the corresponding n-best hidden representations. The MT model uses each of the the n-best source token sequence or their hidden representations to produce the n-best target token sequence. For the given utterance all the hypotheses obtained combined and the likelihood for each hypothesis is computed as shown in equation(\ref{coupling_decoders}) and the best hypothesis is considered as the translation for the given utterance. The hypotheses from ASR and MT  models are produced with a beam size of 10, 5 respectively. The model has two transformer decoders each with a beam search in the pipeline, to reduce the memory and time complexity of the model the vectorized beam search described~\cite{vectorized_beam_search} has been used. The block-diagram describing the proposed model and model ensembling is presented in Figure.~\ref{Multi_task_training_with_ASR_auxilary_loss}. The performance of the Joint SLT systems are in Tables.~\ref{Transformer_Joint_ASR_MT_BPE}~\ref{Transformer_Joint_ASR_MT_CHAR_BPE}. \\

\begin{figure}[!htbp]
	\begin{center}
		\includegraphics[width=80mm,height=70mm]{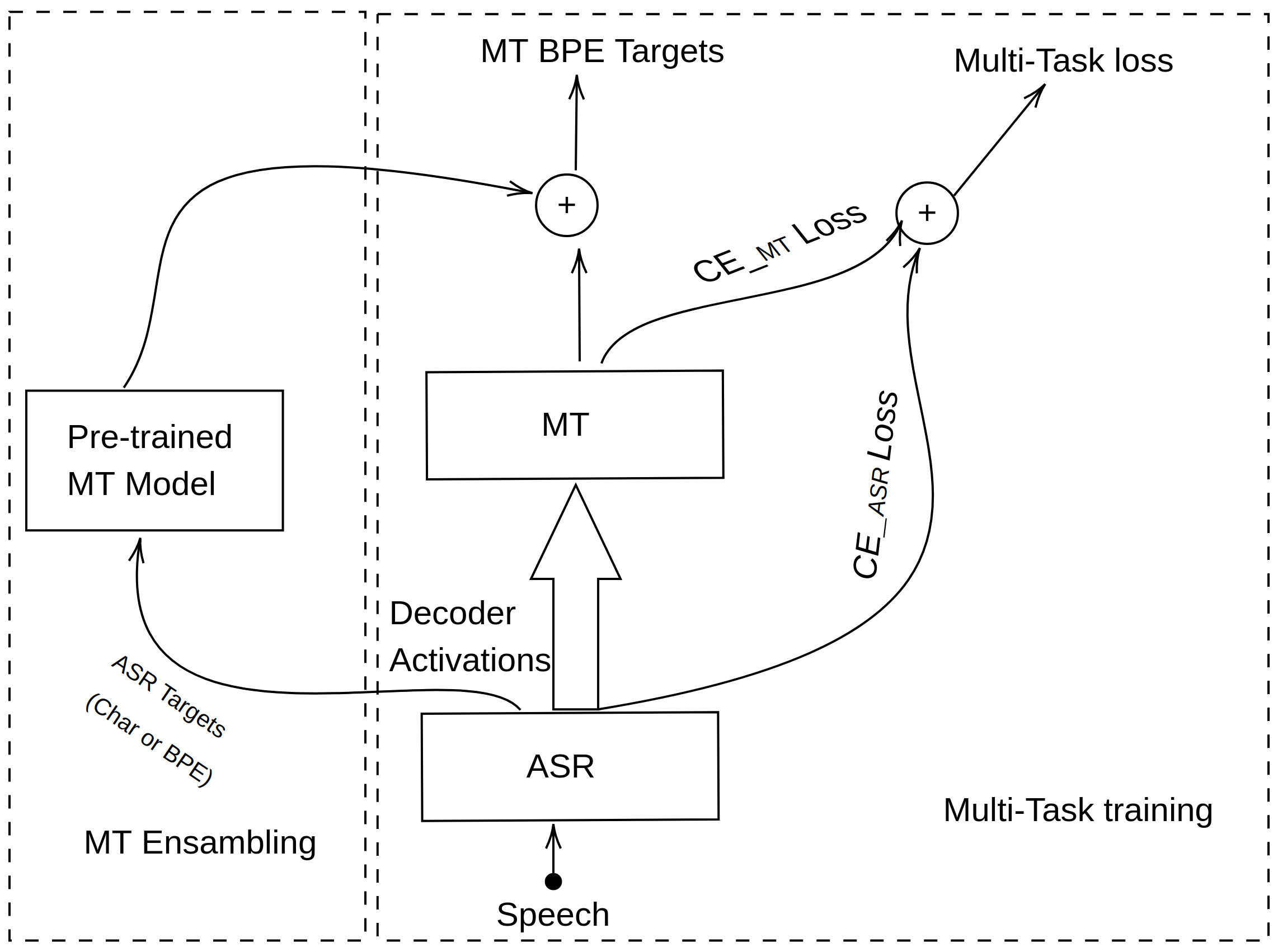}
		\caption{Block diagram describing the Multi-Task training with ASR objective as an auxiliary loss and MT ensembling.}
		\label{Multi_task_training_with_ASR_auxilary_loss}
	\end{center}
\end{figure}
Table~\ref{Transformer_Joint_ASR_MT_BPE},~\ref{Transformer_Joint_ASR_MT_CHAR_BPE} are the performances of Jointly trained cascade SLT systems. The SLT systems reported in Table~\ref{Transformer_Joint_ASR_MT_BPE} uses the ASR with BPE as target units and Table~\ref{Transformer_Joint_ASR_MT_CHAR_BPE} uses characters as target units. Column~1 of Tables~\ref{Transformer_Joint_ASR_MT_BPE},~\ref{Transformer_Joint_ASR_MT_CHAR_BPE} describes the architecture of the systems and Row~1 is the different test sets used for evaluating the systems. The trained SLT systems have two different types of MT modules i.e., Joint-MT and Ext-MT. The MT module trained with discrete symbol sequence as input is termed as Ext-MT. Joint-MT is trained as a part of the joint training and it uses the neural hidden representations from the ASR module as input. Both these models are either independently used or an ensemble of them is used for obtaining the translations. Row~3 of the tables uses the n-best sources tokens from the Ext-ASR and Ext-MT translates the tokens and the best hypothesis is obtained as shown in equation(\ref{coupling_decoders}).
SLT system in Row~4 is uses the source tokens from Ext-ASR to generate neural hidden representations from Joint-ASR and the activations from the pre-softmax layer is used by Joint-MT for generating the translations. The source tokens from Ext-ASR is used as the partial decoded token sequence and an additional token is generated using Joint-ASR which gives the neural hidden representations from Joint-ASR for the corresponding source tokens form the Ext-ASR. Row~5 uses the source tokens and the corresponding neural hidden representations from the Ext-ASR and the corresponding MT modules are used to generate translations and the translations from both the models are ensembled by averaging the softmax distributions and the averaged softmax distribution is used for generating th
e hypotheses. For the SLT systems in Rows~6, 7 and 8 the ASR hypotheses is obtained from the ASR module obtained in joint training~(Joint-ASR). The systems in Row~9, 10, 11 uses the ASR hypotheses obtained by ensembling the Ext.ASR and joint-ASR models. \\

\begin{table*}[!ht]
\renewcommand{\arraystretch}{1}
\setlength{\tabcolsep}{2pt}
\scriptsize
\begin{center}
\caption{Performances of Joint SLT systems under various ensemble combinations. The MT models in this table are trained with BPE as input and output granularity.}
\label{Transformer_Joint_ASR_MT_BPE}
\begin{tabular}{|l|cc|cc|cc|cc|cc|}
\hline
BPE-BPE&\multicolumn{2}{c|}{dev-2010}&\multicolumn{2}{c|}{tst-2010}&\multicolumn{2}{c|}{tst-2013}&\multicolumn{2}{c|}{tst-2014}&\multicolumn{2}{c|}{tst-2015}\\
\hline
\shortstack{Ensemble}& BLEU&WER&BLEU&WER&BLEU&WER&BLEU&WER&BLEU&WER\\
\hline
[Ext-ASR]$\Longrightarrow$[Ext-MT]&19.86&&17.44&&20.21&&17.78&&15.64&~\\

[Ext-ASR]$\Longrightarrow$[Joint-MT]&20.76&{16.33}&19.14&{19.58}&22.30&{17.50}&17.43&{13.54}&16.83&{14.49}\\

[Ext-ASR]$\Longrightarrow$[Joint-MT + Ext-MT]&22.47&&20.00&&23.55&&18.60&&17.51&~\\
\hline
[Joint-ASR]$\Longrightarrow$[Ext-MT]&16.06&&14.97&&15.94&&12.78&&13.79&~\\

[Joint-ASR]$\Longrightarrow$[Joint-MT]&18.44&{34.91}&16.44&{37.87}&17.10&{34.28}&13.53&{34.72}&14.00&{32.62}\\

[Joint-ASR]$\Longrightarrow$[Joint-MT + Ext-MT]&18.72&&16.53&&18.55&&14.42&&14.75&\\
\hline
[Ext-ASR + Joint-ASR]$\Longrightarrow$[Ext-MT]&16.97&&16.00&&18.61&&14.36&&15.04&~\\

[Ext-ASR + Joint-ASR]$\Longrightarrow$[Joint-MT] &19.59&{23.68}&17.63&{26.74}&20.27&{23.72}&15.35&{21.03}&15.59&{22.54}\\

[Ext-ASR + Joint-ASR]$\Longrightarrow$[Joint-MT + Ext-MT]&20.05&&17.89&&21.39&&15.92&&16.86&\\
\hline
\end{tabular}
\end{center}
\end{table*}
\begin{table*}[!ht]
\renewcommand{\arraystretch}{1}
\setlength{\tabcolsep}{2pt}
\scriptsize
\caption{Performances of Joint SLT systems under various ensemble combinations. The MT models in this table are trained with characters-BPE as input-output granularities.}
\label{Transformer_Joint_ASR_MT_CHAR_BPE}
\begin{center}
\begin{tabular}{|l|cc|cc|cc|cc|cc|}
\hline
CHAR-BPE&\multicolumn{2}{c|}{dev-2010}&\multicolumn{2}{c|}{tst-2010}&\multicolumn{2}{c|}{tst-2013}&\multicolumn{2}{c|}{tst-2014}&\multicolumn{2}{c|}{tst-2015}\\
\hline
\shortstack{Ensemble}& BLEU&WER&BLEU&WER&BLEU&WER&BLEU&WER&BLEU&WER\\
\hline
[Ext-ASR]$\Longrightarrow$[Ext-MT]&19.86 && 18.00 && 20.02 && 16.93 && 15.47&~\\

[Ext-ASR]$\Longrightarrow$[Joint-MT]&16.33&{18.40}&19.58&{21.31}&17.50&{18.97}&13.54&{14.06}&14.49&{15.25}\\

[Ext-ASR]$\Longrightarrow$[Joint-MT + Ext-MT]&21.26&&19.10&&21.66&&18.36 &&16.73&~\\
\hline
[Joint-ASR]$\Longrightarrow$[Ext-MT]&16.87  && 14.94  && 14.90  && 12.50  && 12.54 &~\\ 

[Joint-ASR]$\Longrightarrow$[Joint-MT]&15.85&{41.97}&13.73&{45.30}&14.53&{40.14}&11.10&{41.02}&12.48&{39.02}\\

[Joint-ASR]$\Longrightarrow$[Joint-MT + Ext-MT]&17.45  && 15.20  && 15.93  && 13.42  && 13.72 &~\\
\hline
[Ext-ASR + Joint-ASR]$\Longrightarrow$[Ext-MT]&18.25  && 15.57  && 18.81  && 14.54  && 13.93 &~\\

[Ext-ASR + Joint-ASR]$\Longrightarrow$[Joint-MT]&17.49&{31.31}&15.18&{25.24}&18.47&{26.28}&13.71&{26.28}&14.36&{26.53}\\

[Ext-ASR + Joint-ASR]$\Longrightarrow$[Joint-MT + Ext-MT]&19.20  && 16.95  && 19.67  && 15.62  && 15.83 &~\\
\hline
\end{tabular}
\end{center}
\end{table*}

From Table~\ref{Transformer_Joint_ASR_MT_BPE},~\ref{Transformer_Joint_ASR_MT_CHAR_BPE}, it can be observed that the better ASR performance has always improved the Translation performance. SLT systems which passes neural hidden representations between ASR and MT have performed better than the models which passed token sequences, this difference can be seen by comparing rows~3,4; 6,7; and 8,9 in Table~\ref{Transformer_Joint_ASR_MT_BPE},~\ref{Transformer_Joint_ASR_MT_CHAR_BPE}. Ensembling the Joint.MT models with the MT models trained for discrete input sequences~(Ext.MT) has improved the performances. The Jointly trained ASR and MT models have better performance with the ASR output is BPE's, the joint optimization has better trained the ASR model when BPE's are targets rather than using the characters as targets, this can be observed by comparing rows~7 in Table~\ref{Transformer_Joint_ASR_MT_BPE}, ~\ref{Transformer_Joint_ASR_MT_CHAR_BPE}. All the models ensembled in this work are ensembles by averaging the Softmax distributions, having a additional hyper parameter while averaging could improve the performances.
\newpage
\section{Related Work}

Some of the works related to the proposed model are described briefly: Initial SLT systems are cascaded systems where the pipeline consists of an ASR systems followed by an MT system~\cite{Ney_cascade_SLT,ney_coupling_recognitin_translation}~\cite{attention_passing_models}. Recently End-to-End models have been used for training SLT systems~\cite{neural_MT},~\cite{Listen_and_translate},~\cite{Attention_translation_without_transcription}. Transformer models are used for training End-to-End SLT systems in ~\cite{SPMT_tranformer_diag_loss,Espnet_ST_toolkit} and initializing the weights from pre-trained ASR encoder and MT Decoder has improved the performance of these models significantly~\cite{Espnet_ST_toolkit}. Multi-lingual speech Translation models has been studied in~\cite{one_to_many_ST,Must_c_dataset} ~\cite{inaguma2019multilingual}. After pre-training the ASR and MT models only parts of them are used in End-to-End speech translation and they are fine-tuned for the task. To have the whole structure of the network intact in pre-training and fine-tuning, a speech translation model is designed by stacking an ASR encoder trained using CTC and an MT model~\cite{bridging_pretraining_finetuning}. The text sequence is synthetically modified to match the CTC-ASR token sequence while training the MT model. Similar structure has been optimized with an additional auxiliary loss~(i.e., L1-distance between the hidden representations of MT encoder and pre-trained BERT model) in~\cite{Triple_Supervision}. A curriculum based approach is followed in~\cite{wang_curriculum}, to train the encoder of and End-to-End ST model to progressively learn to concepts of transcription of an audio and syntax of the source language and trasltion to the target language. Cascade ASR + MT in~\cite{wang_curriculum}, is trained with additional data for both ASR and MT. A similar framework has been used ~\cite{Qianqian_SDST}, rather that inserting additional blanks, an acoustic unit shrinking layer is used to generate phoneme level labels from frame level labels and the corresponding representations are used to train the MT models. The MT decoder is initialized from a pre-trained model for better performance. SLT systems in various data environments are trained spec-augment have shown higher performances in~\cite{bahar_SLT_SpecAug}. The capability of the model to translate all the n-best output sequences and the MT model being able to choose best among all the translations gives better coupling and additional capacity to the model. The proposed model is similar to the model described in~\cite{sperber2019attention}, where the context vectors from the ASR decoder are passed as input to the MT model. During the inference, the proposed model is similar to~\cite{NMT_coupled_decoding}, where back translation likelihoods are used to re-rank the hypotheses, proposed model uses the likelihoods from ASR decoder to re-rank the MT hypotheses. 
\begin{table*}[!ht]
\renewcommand{\arraystretch}{1}
\setlength{\tabcolsep}{2pt}
\scriptsize
\begin{center}
\caption{Comparing the performance of the proposed model with the other state-of-the-art models.}
\label{Transformer_ASR}
\begin{tabular}{lccccc}
\hline
ASR-system &dev-2010 & tst2010 & tst-2013 & tst-2014 & tst-2015\\
\hline
Hirofumi Inaguma.et al~\cite{inaguma2019multilingual}-End-to-End &-&-&14.6&-&-\\
Chengyi Wang.et al(2020)~\cite{bridging_pretraining_finetuning} - End-to-End&-&17.61&17.67&15.73&14.94\\
Parnia Bahar~\cite{bahar_SLT_SpecAug}-End-to-End &21.3&-&-&-&20.9\\
Qianqian Dong~\cite{Triple_Supervision}&-&16.43&15.61&13.77&15.29\\
Chengyi Wang.et al(2020)~\cite{wang_curriculum}-End-to-End &-&-&18.15&-&-\\
Chengyi Wang.et al(2020)~\cite{wang_curriculum}-Cascade &-&-&22.16&-&-\\
Chengyi Wang.et al(2020)~\cite{bridging_pretraining_finetuning} - Cascade&-&13.38&15.84&12.94&13.79\\
Chengyi Wang.et al(2020)~\cite{bridging_pretraining_finetuning} - Cascade + reseg &-&17.12&17.77&14.94&15.01\\
Qianqian Dong.et al(2020)~\cite{Qianqian_SDST} &-&21.31 & 18.63 & 16.20 & 17.72\\
Parnia Bahar~\cite{bahar_SLT_SpecAug}-Cascade &24.7&-&-&-&24.4\\
\hline
Proposed model&22.47&20.00&23.55&18.60&17.51\\
\hline
\end{tabular}
\end{center}
\end{table*}

\newpage
\bibliographystyle{IEEEbib}
\bibliography{SP_MT.bib}
\end{document}